\documentclass[runningheads]{llncs}
\PassOptionsToPackage{colorlinks=true,linkcolor=blue,citecolor=blue,urlcolor=blue}{hyperref}
\usepackage[T1]{fontenc}
\usepackage{amsmath}
\usepackage{amssymb}
\usepackage{bm}
\usepackage{graphicx,verbatim}
\usepackage{booktabs}
\usepackage{multirow}
\usepackage{orcidlink} 
\usepackage{stfloats}
\usepackage{graphicx}
\usepackage{xcolor}
\usepackage{colortbl}
\usepackage{booktabs}
\usepackage{multirow}
\usepackage{array}

\definecolor{bestblue}{HTML}{DEEBF7}      
\definecolor{secondgray}{HTML}{F0F0F0}    
\definecolor{headergray}{HTML}{D9D9D9}    
\definecolor{dicemint}{HTML}{FAE3D5}      
\definecolor{ravdpeach}{HTML}{E5F0E8}     
\definecolor{asdlavender}{HTML}{E6E0F5}   
\definecolor{hd95cyan}{HTML}{E5F6FD}      
\definecolor{deltagreen}{HTML}{199935}    
\begin{document}
\title{Mutual Distillation of Dual-Foundation Models for Semi-Supervised PET/CT Segmentation}
%
%
\author{
Fuyou Mao\inst{1}\textsuperscript{\textdagger}\orcidlink{0009-0001-1955-3110} \and
Beining Wu\inst{2}\textsuperscript{\textdagger}\orcidlink{0009-0006-0317-0920} \and
Yanfeng Jiang\inst{3}\textsuperscript{\textdagger}\orcidlink{0009-0000-6229-7477} \and
Bohan Xu\inst{4}\orcidlink{0009-0002-7479-4045} \and
Lixin Lin\inst{1}\orcidlink{0009-0009-9076-5114} \and
Naye Ji\inst{3}\textsuperscript{*}\orcidlink{0000-0002-6986-3766} \and
Hao Zhang\inst{1}\textsuperscript{*}\orcidlink{0009-0002-4983-0841} \and
Yan Tang\inst{1}\orcidlink{0009-0006-5335-4306}
}

\authorrunning{F. Mao et al.}

\institute{
Central South University, Changsha 410083, China \and
Hangzhou Dianzi University, Hangzhou 310018, China \and
Communication University of Zhejiang, Hangzhou 310018, China \and
Northeastern University, Shenyang 110819, China \\[0.3em]
\email{jinaye@cuz.edu.cn}, \email{hao@csu.edu.cn}
}

\maketitle

\bgroup
\footnotesize
\renewcommand{\thefootnote}{}
\footnotetext{\textsuperscript{\textdagger} Equal Contribution.}
\footnotetext{\textsuperscript{*} Corresponding authors.}
\egroup
\begin{abstract}
Organ segmentation from PET/CT is critical for quantitative analysis and radiotherapy planning in oncology.
To ease the high annotation cost of PET/CT segmentation, semi-supervised learning (SSL) provides a practical and effective solution for developing deep models with limited labeled data.
Recent developments in visual foundation models have demonstrated remarkable adaptability with improved efficiency.
In this work, we propose a mutual distillation framework that seamlessly exploits both structural and functional foundation models, which act as modality-specific generalists for distilling knowledge from structural CT and metabolic PET imaging.
By bridging the gap between the task-specific precision of student models and the segmentation priors of generalist foundation models, we propose \textbf{MuDuo}, a mutual distillation framework that synergistically leverages SAM-Med3D for CT and SegAnyPET for PET to distill their knowledge into a lightweight student network. Our approach eliminates the need for manual prompts while maximizing the utility of unlabeled data for automatic segmentation, achieving state-of-the-art performance on the AutoPET dataset with only 5 labeled cases. Our source code is available at \href{https://github.com/Wu-beining/MuDuo}{https://github.com/Wu-beining/MuDuo}.

\end{abstract}
\section{Introduction}
Accurate and automated organ segmentation in whole-body PET/CT is a critical prerequisite for quantitative metabolic analysis, precise disease staging, and targeted radiotherapy planning \cite{wang2025robust}.
In clinical practice, paired PET/CT scans inherently provide complementary structural and functional information \cite{godinez2025total}.
The primary difficulty lies in the anatomical-functional mismatch. PET imaging is characterized by high sensitivity to metabolic activity, which frequently results in blurred boundaries and spill-over artifacts. Consequently, resolving this spatial discrepancy to achieve efficient and accurate cross-modal information fusion remains a formidable challenge \cite{qin2025multimodal}.
Another challenge is the scarcity of high-quality pixel-level 3D annotations, which requires synchronized expertise from both radiologists and nuclear medicine specialists to ensure anatomical precision and metabolic accuracy \cite{tajbakhsh2020embracing}. 

To address data scarcity, semi-supervised learning (SSL) mitigates this burden by leveraging both limited labeled and abundant unlabeled data for training \cite{jiao2024learning}. Traditional SSL methods primarily follow two paradigms: pseudo-labeling and consistency regularization. Pseudo-labeling methods generate artificial supervisory signals by utilizing the model's own high-confidence predictions on unlabeled data, iteratively expanding the effective training set \cite{lee2013pseudo,iscen2019label}. However, these approaches often suffer from confirmation bias, where erroneous predictions are reinforced because the model is encouraged to maintain invariance under different perturbations, leveraging the manifold assumption that similar inputs should yield consistent outputs \cite{laine2017temporal,tarvainen2017mean}. While effective, such methods typically rely solely on the student model's own predictions without exploiting external knowledge sources, limiting their performance ceiling in complex medical imaging scenarios.
To further push the boundaries of SSL, recent efforts have turned to utilize foundation models. 
A leading strategy is to integrate the Segment Anything Model (SAM) \cite{kirillov2023segment} and its medical variants to utilize its powerful segmentation priors to enhance semi-supervised medical image segmentation. Representative frameworks such as SemiSAM \cite{SemiSAM} has introduced a compelling specialist-generalist paradigm, where a lightweight student model generates initial prompts and a frozen SAM then refines these prompts into high-quality masks. This approach allows models to distillate large-scale pre-trained knowledge to generalize from minimal labeled data, thereby significantly reducing the dependency on extensive manual annotations.  Despite yielding promising results, they are primarily designed for single-modality scenarios and fail to capture the deep semantic interdependence between structural and functional imaging for PET/CT segmentation, which inherently requires the synergistic integration of anatomical context from CT and metabolic activity from PET to achieve accurate organ delineation.

To bridge this gap, we propose a novel mutual distillation framework that synergistically exploits structural and functional foundation models for semi-supervised PET/CT organ segmentation. Unlike existing single modality approaches, our approach utilizes two complementary foundation models including SAM-Med3D \cite{wang2024sammed3dgeneralpurposesegmentationmodels} for CT and SegAnyPET \cite{zhang2025seganypet} for PET to mutually distill knowledge to a lightweight student network. 
Specifically, our framework operates two distinct pathways: 1) a \textit{prompt sampling} branch, where the student model's predictions serve as mask prompts to guide SAM-Med3D and SegAnyPET to leverage the segmentation priors for pseudo label generation; and 2) a \textit{consistency regularization} branch, where generated pseudo labels from SAM-Med3D and SegAnyPET are utilized for regularization to distill the knowledge of foundation models.
To ensure high-quality pseudo supervision, we propose an IoU-based consensus filtering mechanism that retains only the top 50\% most reliable predictions where both modalities agree. These selected pseudo-labels are fused and used to supervise the student, enabling effective knowledge transfer from unlabeled data without manual prompting. This mutual distillation strategy seamlessly integrates structural and functional segmentation priors, eliminating the need for expensive dual-modality annotations while maximizing the utility of unlabeled PET/CT volumes.

\section{Method}

\subsection{Problem Formulation}

Mathematically, we define the 3D input images as $\boldsymbol{X}_{c}\in\mathbb{R}^{W\times H\times D}$ for CT imaging and $\boldsymbol{X}_{p}\in\mathbb{R}^{W\times H\times D}$ for PET imaging, which are jointly fed into the student network. The goal of semi-supervised medical image segmentation is to predict the per-voxel label map $\widehat{\boldsymbol{Y}}\in\{0,1,\ldots,C\}^{W\times H\times D}$, where $C$ is the number of classes. Our training set $\mathcal{D}$ consists of $N$ labeled data and $M$ unlabeled data (generally $N\ll M$), expressed as two subsets: $\boldsymbol{\mathcal{D}}=\boldsymbol{\mathcal{D}}^{l}\cup\boldsymbol{\mathcal{D}}^{u}$, where $\boldsymbol{\mathcal{D}}^{l}=\{\boldsymbol{X}_{c}^{l},\boldsymbol{X}_{p}^{l},\boldsymbol{Y}\}$ represents the labeled set with corresponding annotation $\textbf{Y}$, and $\boldsymbol{\mathcal{D}}^{u}=\{\boldsymbol{X}_{c}^{u},\boldsymbol{X}_{p}^{u}\}$ represents the unlabeled set without the annotation mask.

\begin{figure*}[tb]
    \centering
    \includegraphics[width=0.9\linewidth]{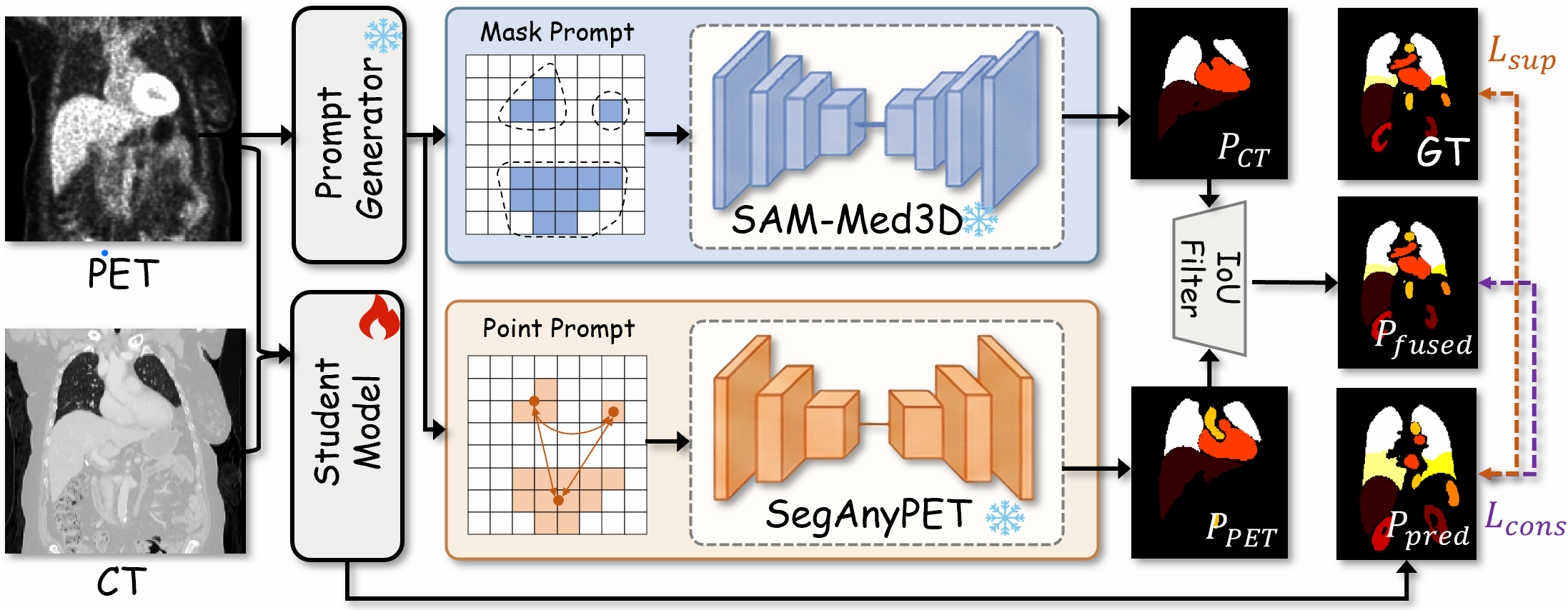}
    \caption{\textbf{Overview of the proposed MuDuo framework.} The student model processes dual-modal PET/CT inputs. For unlabeled data, we generate pseudo labels through two pathways: 1) CT branch: student predictions serve as mask prompts for SAM-Med3D; 2) PET branch: the prompt generator converts student features to point prompts for SegAnyPET. Predictions from both branches are fused and filtered based on IoU, with the resulting pseudo labels supervising the student via consistency loss.}
    \label{fig:framework}
\end{figure*}

\subsection{Framework Overview}

As illustrated in Fig. \ref{fig:framework}, \textbf{MuDuo} comprises three main components: (1) a learnable student U-Net $\boldsymbol{\mathcal{S}}_{\theta}$ that processes dual-modal inputs; (2) two frozen foundation model teachers specialized for CT ($\boldsymbol{\mathcal{T}}_{\text{CT}}$: SAM-Med3D) and PET ($\boldsymbol{\mathcal{T}}_{\text{PET}}$: SegAnyPET); and (3) a lightweight prompt generator $\boldsymbol{\mathcal{G}}_{\phi}$ (UNet) that bridges the student and PET teacher.

\subsection{Pseudo Annotation Generation with Foundation Models}
For unlabeled data, we leverage a pretrained prompt generator $\mathcal{G}_\phi$ to generate pseudo-labels through dual complementary pathways, each adapted to the unique characteristics of CT and PET modalities.

\textbf{CT Pathway.} We employ the predictions of the pretrained prompt generator $\mathcal{G}_\phi$ on CT data as mask prompts for SAM-Med3D\cite{wang2024sammed3dgeneralpurposesegmentationmodels}. For each class $c \in \{1, \ldots, C-1\}$, we extract the probability map $\mathbf{P}_{\text{init}}^c \in \mathbb{R}^{1\times D\times H\times W}$ from $\mathcal{G}_\phi$ and use it as the mask prompt:
\begin{equation}
\label{eq:ct_pathway}
\begin{aligned}
\boldsymbol{\mathcal{P}}_{\text{CT}}^{c} &= \sigma\left(\boldsymbol{\mathcal{T}}_{\text{CT}}\left(\boldsymbol{X}_{c}; \text{mask}=\text{downsample}\left(\boldsymbol{\mathcal{P}}_{\text{init}}^{c}\right)\right)\right),
\end{aligned}
\end{equation}
where $\sigma(\cdot)$ denotes the sigmoid function, and the mask is downsampled to match SAM's latent resolution (32$\times$ reduction). SAM-Med3D refines the initial predictions based on CT anatomical priors.

\textbf{PET Pathway.} We utilize the same prompt generator $\mathcal{G}_\phi$ to generate point prompts. Given PET input $\mathbf{X}_p$, the model produces coarse logits $\mathbf{Z}_p = \mathcal{G}_\phi(\mathbf{X}_p)$. For class $c$, we obtain a binary mask $\mathbf{M}_p^c = \mathbb{I}[\arg\max(\mathbf{Z}_p) = c]$.

Instead of relying on a single centroid, we propose multi-point sampling to better capture the spatial extent of each organ. We apply light Gaussian smoothing to $\mathbf{M}_p^c$ and sample the top-$K$ highest probability locations (default $K=3$):
\begin{equation}
\label{eq:multipoint}
\begin{aligned}
\left\{\left(z_{i}, y_{i}, x_{i}\right)\right\}_{i=1}^{K} &= \text{top-}K\left(\boldsymbol{\mathcal{M}}_{p}^{c} * \boldsymbol{\mathcal{G}}_{\sigma}\right),
\end{aligned}
\end{equation}
where $*$ denotes convolution. These points, along with foreground labels $\mathbf{L} \in \mathbb{R}^{K}$, prompt SegAnyPET:
\begin{equation}
\label{eq:pet_pathway}
\begin{aligned}
\boldsymbol{\mathcal{P}}_{\text{PET}}^{c} &= \sigma\left(\boldsymbol{\mathcal{T}}_{\text{PET}}\left(\boldsymbol{X}_{p}; \text{points}=\left\{\left(z_{i}, y_{i}, x_{i}\right)\right\}, \text{labels}=\boldsymbol{L}\right)\right),
\end{aligned}
\end{equation}

Multi-point prompting is crucial for PET imaging where organs often exhibit heterogeneous uptake, and single points may miss peripheral regions or capture only the most active portion.

\subsection{Mutual Distillation with MuDuo}

\subsubsection{IoU-Based Quality Filtering}
Not all pseudo labels are equally reliable. We propose an IoU-based filtering mechanism to select high-quality predictions for distillation. For each class $c$ and sample $i$, we compute the intersection-over-union between CT and PET branch predictions:
\begin{equation}
\label{eq:iou}
\begin{aligned}
\text{IoU}_{i}^{c} &= \frac{\sum\left(\boldsymbol{\mathcal{P}}_{\text{CT},i}^{c} \odot \boldsymbol{\mathcal{P}}_{\text{PET},i}^{c}\right)}{\sum\left(\boldsymbol{\mathcal{P}}_{\text{CT},i}^{c} + \boldsymbol{\mathcal{P}}_{\text{PET},i}^{c} - \boldsymbol{\mathcal{P}}_{\text{CT},i}^{c} \odot \boldsymbol{\mathcal{P}}_{\text{PET},i}^{c}\right) + \boldsymbol{\epsilon}},
\end{aligned}
\end{equation}
where $\odot$ denotes element-wise multiplication and $\epsilon=10^{-8}$ ensures numerical stability. High IoU indicates consensus between CT and PET modalities, suggesting reliable predictions.

For each class, we retain only the top 50\% samples with highest IoU scores:
\begin{equation}
\label{eq:filtering}
\begin{aligned}
\boldsymbol{\mathcal{I}}_{c} &= \text{top-}50\%\left(\left\{\text{IoU}_{i}^{c}\right\}_{i=1}^{B}\right),
\end{aligned}
\end{equation}
where $B$ is the batch size. Samples with invalid prompts are assigned IoU $=-1$ to ensure exclusion.

\subsubsection{Pseudo Label Fusion and Consistency Loss}
Inspired by the dynamically mixed pseudo labeling strategy~\cite{luo2022scribbleseg}, we generate hard pseudo labels by randomly mixing predictions from CT and PET experts:

\begin{equation}
\label{eq:dynamic_mix}
\begin{aligned}
\boldsymbol{\mathcal{P}}_{\text{fused}}^{c} &= \arg\max\left[\boldsymbol{\alpha} \cdot \boldsymbol{\mathcal{P}}_{\text{CT}}^{c} + \left(1-\boldsymbol{\alpha}\right) \cdot \boldsymbol{\mathcal{P}}_{\text{PET}}^{c}\right], \quad \boldsymbol{\alpha} \sim \boldsymbol{\mathcal{U}}(0,1),
\end{aligned}
\end{equation}

where $\alpha$ is randomly sampled from a  uniform distribution $\mathcal{U}(0,1)$ at each iteration for every sample. This dynamic mixing strategy introduces perturbation into the pseudo label generation process, which helps prevent the model from memorizing its own predictions and enhances the diversity of supervision signals. The $\arg\max$ operation converts the mixed soft probabilities into hard pseudo labels, providing more decisive supervision compared to soft averaging.

The consistency loss encourages the student to match these high-quality pseudo labels using mean squared error:
\begin{equation}
\label{eq:consistency}
\begin{aligned}
\boldsymbol{\mathcal{L}}_{\text{cons}} &= \frac{1}{\left|\boldsymbol{\mathcal{C}}\right|}\sum_{c \in \boldsymbol{\mathcal{C}}} \frac{1}{\left|\boldsymbol{\mathcal{I}}_{c}\right|}\sum_{i \in \boldsymbol{\mathcal{I}}_{c}} \left\|\boldsymbol{\mathcal{P}}_{s,i}^{c} - \boldsymbol{\mathcal{P}}_{\text{fused},i}^{c}\right\|_{2}^{2},
\end{aligned}
\end{equation}
where $\mathcal{C}$ is the set of classes present in the batch. This formulation ensures that the student learns only from reliable, consensus-based pseudo annotations.

\subsubsection{Gaussian Ramp-Up Weight Scheduling}
Following common practice in SSL \cite{tarvainen2017mean}, we apply a gaussian ramp-up to the consistency weight:
\begin{equation}
\label{eq:rampup}
\begin{aligned}
\boldsymbol{\lambda}(t) &= \boldsymbol{\lambda}_{\text{max}} \cdot \exp\left(-5\left(1 - \frac{t}{t_{\text{ramp}}}\right)^{2}\right),
\end{aligned}
\end{equation}
where $t$ is the current iteration and $t_{\text{ramp}}$ is the ramp-up period (set to 200 epochs). This prevents the student from being overwhelmed by noisy pseudo labels in early training.

\section{Experiments}
\begin{figure*}[t]
    \centering
    \includegraphics[width=1.0\linewidth]{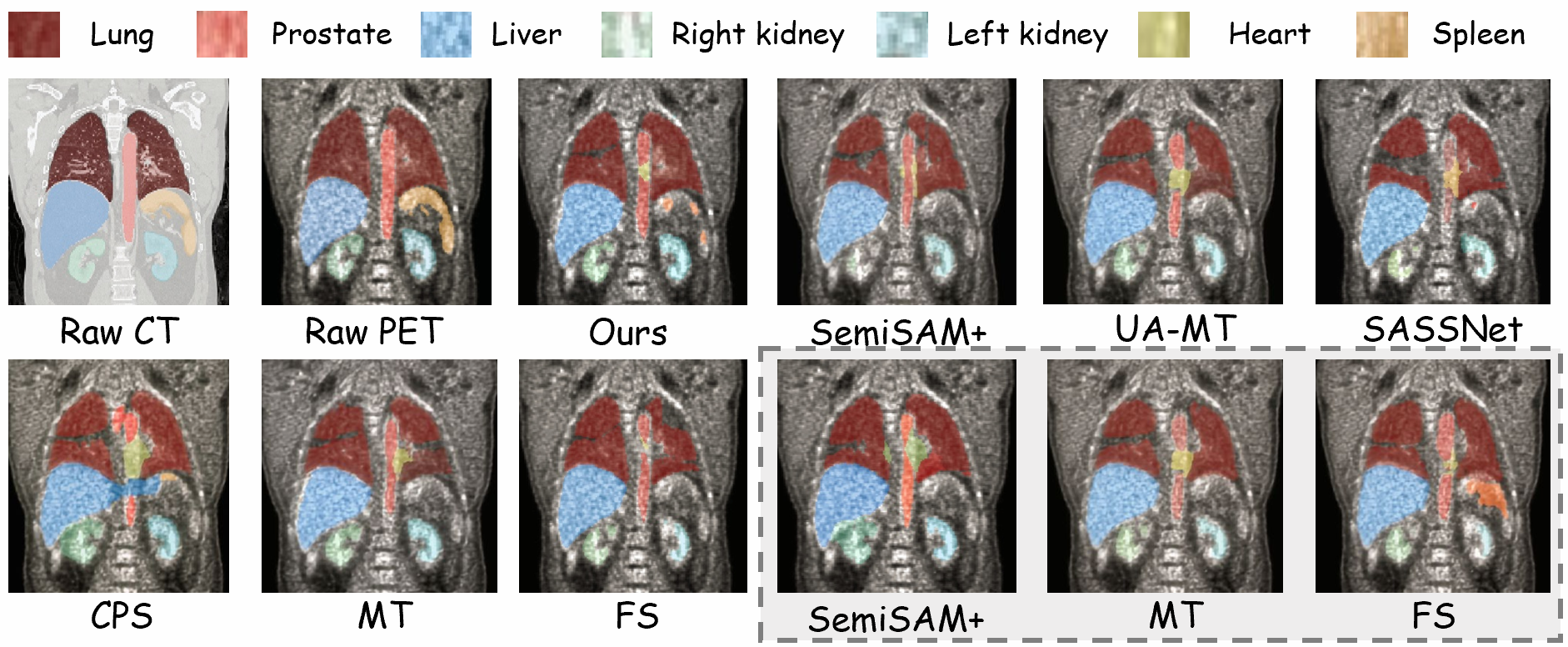}
    \caption{\textbf{Qualitative comparison of multi-organ segmentation on PET/CT across methods.} The top row shows segmentation results on a coronal view, and the bottom row shows additional visualizations including PET-only semi-supervised methods (SemiSAM+, MT, FS in dashed box).}
    \label{fig2}
\end{figure*}

\subsection{Dataset and Preprocessing}

\begin{table*}[p]
\centering
\caption{Quantitative comparison of PET organ segmentation performance across different settings. The L/U column indicates the number of labeled and unlabeled cases. Metrics include Dice (\%), RAVD (\%), ASD (voxel), and 95HD (voxel). The best results are highlighted in \colorbox{bestblue}{blue} and the second-best results are highlighted in \colorbox{secondgray}{gray}.}
\label{tab:segmentation_results}
\resizebox{\textwidth}{!}{%
\begin{tabular}{l!{\vrule width 1.5pt}l!{\vrule width 1.0pt}c!{\vrule width 1.0pt}cccc}
\toprule[1.5pt]
\textbf{Modality} & \textbf{Method} & \textbf{L/U} & \cellcolor{dicemint}\textbf{Dice[\%] $\uparrow$} & \cellcolor{ravdpeach}\textbf{RAVD[\%] $\downarrow$} & \cellcolor{asdlavender}\textbf{ASD[voxel] $\downarrow$} & \cellcolor{hd95cyan}\textbf{HD95[voxel] $\downarrow$} \\
\midrule[1.5pt]
\rowcolor{headergray}
\multicolumn{7}{l}{\textbf{Setting: 5 Labeled Cases}} \\
\midrule[1.5pt]
\multirow{3}{*}{PET Only} 
& Fully-Supervised & $5 / 0$   & $0.3186_{\pm 0.1849}$ & $0.4108_{\pm 0.2050}$ & $51.14_{\pm 33.50}$ & $20.18_{\pm 14.63}$ \\
& MT~\cite{tarvainen2017mean} & $5 / 953$ & $0.3644_{\pm 0.2015}$ & $0.3521_{\pm 0.2451}$ & $34.17_{\pm 19.54}$ & $17.21_{\pm 10.27}$ \\
& SemiSAM+~\cite{zhang2025semisam} & $5 / 953$ & $0.2997_{\pm 0.1794}$ & $0.5587_{\pm 0.2008}$ & $40.60_{\pm 27.61}$ & $14.93_{\pm 12.53}$ \\
\midrule[1.0pt]
\multirow{7}{*}{PET/CT} 
& Fully-Supervised & $5 / 0$   & $0.3852_{\pm 0.2471}$ & $0.3376_{\pm 0.2272}$ & $36.35_{\pm 18.22}$ & $15.06_{\pm 9.91}$ \\
& MT~\cite{tarvainen2017mean} & $5 / 953$ & $0.3896_{\pm 0.2460}$ & $0.3626_{\pm 0.2089}$ & $46.94_{\pm 29.09}$ & $17.60_{\pm 11.32}$ \\
& SASSNet~\cite{li2020shape} & $5 / 953$ & $0.4073_{\pm 0.2619}$ & \cellcolor{bestblue}$0.3263_{\pm 0.2248}$ & $35.29_{\pm 19.41}$ & $15.32_{\pm 10.98}$ \\
& UA-MT~\cite{yu2019uncertainty} & $5 / 953$ & $0.3799_{\pm 0.2246}$ & $0.3438_{\pm 0.1782}$ & $49.59_{\pm 37.98}$ & $17.51_{\pm 10.79}$ \\
& CPS~\cite{chen2021semi} & $5 / 953$ & \cellcolor{secondgray}$0.4148_{\pm 0.2287}$ & \cellcolor{secondgray}$0.3387_{\pm 0.1819}$ & $49.09_{\pm 43.32}$ & $16.88_{\pm 11.36}$ \\
& SemiSAM+~\cite{zhang2025semisam} & $5 / 953$ & $0.3215_{\pm 0.1821}$ & $0.4072_{\pm 0.1740}$ & \cellcolor{secondgray}$31.93_{\pm 19.93}$ & \cellcolor{secondgray}$14.41_{\pm 13.20}$ \\
& \textbf{Ours}    & $\mathbf{5 / 953}$ & \cellcolor{bestblue}$\mathbf{0.4693_{\pm 0.2152}}$ & $0.3704_{\pm 0.1879}$ & \cellcolor{bestblue}$\mathbf{23.65_{\pm 16.60}}$ & \cellcolor{bestblue}$\mathbf{8.50_{\pm 7.25}}$ \\
\midrule[1.5pt]
\rowcolor{headergray}
\multicolumn{7}{l}{\textbf{Setting: 10 Labeled Cases}} \\
\midrule[1.5pt]
\multirow{3}{*}{PET Only} 
& Fully-Supervised & $10 / 0$   & $0.3660_{\pm 0.1949}$ & $0.3475_{\pm 0.2428}$ & $37.53_{\pm 24.36}$ & $15.86_{\pm 14.07}$ \\
& MT~\cite{tarvainen2017mean} & $10 / 948$ & $0.3728_{\pm 0.1934}$ & \cellcolor{bestblue}$0.3383_{\pm 0.2220}$ & $37.21_{\pm 23.84}$ & $15.88_{\pm 12.53}$ \\
& SemiSAM+~\cite{zhang2025semisam} & $10 / 948$ & $0.3108_{\pm 0.1580}$ & $0.4983_{\pm 0.1965}$ & $34.18_{\pm 18.71}$ & $13.47_{\pm 10.49}$ \\
\midrule[1.0pt]
\multirow{7}{*}{PET/CT} 
& Fully-Supervised & $10 / 0$   & \cellcolor{secondgray}$0.4011_{\pm 0.1936}$ & \cellcolor{secondgray}$0.3439_{\pm 0.2179}$ & $27.75_{\pm 17.07}$ & $11.35_{\pm 9.10}$ \\
& MT~\cite{tarvainen2017mean} & $10 / 948$ & $0.3872_{\pm 0.1751}$ & $0.4099_{\pm 0.1924}$ & \cellcolor{secondgray}$25.98_{\pm 13.12}$ & $10.40_{\pm 7.61}$ \\
& SASSNet~\cite{li2020shape} & $10 / 948$ & $0.3604_{\pm 0.1803}$ & $0.4243_{\pm 0.2103}$ & $37.33_{\pm 19.30}$ & $14.31_{\pm 9.27}$ \\
& UA-MT~\cite{yu2019uncertainty} & $10 / 948$ & $0.3809_{\pm 0.1688}$ & $0.4344_{\pm 0.1356}$ & $26.65_{\pm 14.88}$ & \cellcolor{secondgray}$10.35_{\pm 7.63}$ \\
& CPS~\cite{chen2021semi} & $10 / 948$ & $0.3872_{\pm 0.1927}$ & $0.3899_{\pm 0.1711}$ & $28.92_{\pm 13.76}$ & $13.29_{\pm 9.64}$ \\
& SemiSAM+~\cite{zhang2025semisam} & $10 / 948$ & $0.3369_{\pm 0.1682}$ & $0.4839_{\pm 0.1380}$ & $29.38_{\pm 15.85}$ & $11.39_{\pm 8.06}$ \\
& \textbf{Ours}    & $\mathbf{10 / 948}$ & \cellcolor{bestblue}$\mathbf{0.4912_{\pm 0.2196}}$ & $0.3498_{\pm 0.1939}$ & \cellcolor{bestblue}$\mathbf{24.11_{\pm 16.35}}$ & \cellcolor{bestblue}$\mathbf{8.78_{\pm 7.15}}$ \\
\midrule[1.5pt]
\rowcolor{headergray}
\multicolumn{7}{l}{\textbf{Setting: 20 Labeled Cases}} \\
\midrule[1.5pt]
\multirow{3}{*}{PET Only} 
& Fully-Supervised & $20 / 0$   & $0.3832_{\pm 0.1962}$ & $0.3832_{\pm 0.2471}$ & $32.66_{\pm 21.56}$ & $13.99_{\pm 14.11}$ \\
& MT~\cite{tarvainen2017mean} & $20 / 938$ & $0.3893_{\pm 0.2038}$ & $0.3888_{\pm 0.1993}$ & $33.92_{\pm 23.90}$ & \cellcolor{secondgray}$13.92_{\pm 13.59}$ \\
& SemiSAM+~\cite{zhang2025semisam} & $20 / 938$ & $0.3770_{\pm 0.1787}$ & $0.4808_{\pm 0.1708}$ & $27.78_{\pm 13.64}$ & $10.11_{\pm 6.24}$ \\
\midrule[1.0pt]
\multirow{7}{*}{PET/CT} 
& Fully-Supervised & $20 / 0$   & $0.3844_{\pm 0.2120}$ & $0.4344_{\pm 0.1832}$ & $21.81_{\pm 13.16}$ & \cellcolor{secondgray}$8.64_{\pm 7.15}$ \\
& MT~\cite{tarvainen2017mean} & $20 / 938$ & $0.3902_{\pm 0.1930}$ & $0.4282_{\pm 0.1562}$ & $23.01_{\pm 13.91}$ & $8.65_{\pm 7.08}$ \\
& CPS~\cite{chen2021semi} & $20 / 938$ & $0.3928_{\pm 0.1940}$ & \cellcolor{bestblue}$0.3055_{\pm 0.2057}$ & $39.96_{\pm 21.91}$ & $16.21_{\pm 12.04}$ \\
& SASSNet~\cite{li2020shape} & $20 / 938$ & \cellcolor{secondgray}$0.4261_{\pm 0.2249}$ & $0.3563_{\pm 0.1710}$ & \cellcolor{bestblue}$11.44_{\pm 9.10}$ & $27.98_{\pm 18.02}$ \\
& UA-MT~\cite{yu2019uncertainty} & $20 / 938$ & $0.3620_{\pm 0.1710}$ & $0.4355_{\pm 0.1220}$ & $26.09_{\pm 13.81}$ & $9.79_{\pm 6.74}$ \\
& SemiSAM+~\cite{zhang2025semisam} & $20 / 938$ & $0.3811_{\pm 0.1872}$ & $0.4338_{\pm 0.1462}$ & $24.10_{\pm 14.83}$ & $9.06_{\pm 7.72}$ \\
& \textbf{Ours}    & $\mathbf{20 / 938}$ & \cellcolor{bestblue}$\mathbf{0.4956_{\pm 0.2167}}$ & \cellcolor{secondgray}$0.3317_{\pm 0.2114}$ & \cellcolor{secondgray}$19.21_{\pm 16.01}$ & \cellcolor{bestblue}$\mathbf{8.45_{\pm 7.26}}$ \\
\bottomrule[1.5pt]
\end{tabular}%
}
\end{table*}
The proposed framework is evaluated on the AutoPET dataset, which comprises $1,038$ multi-modal 3D PET/CT studies \cite{gatidis2022whole}. For organ-specific segmentation, we utilize a subset of $100$ cases containing expert annotations provided by SegAnyPET \cite{zhang2025seganypet}. We allocate $N$ cases for the training set and the remaining $100-N$ cases for the test set. To assess the performance of the model under varying levels of data scarcity, the training set is divided into multiple labeled settings with $N \in \{5, 10, 20\}$ annotated samples, while the remaining studies from the AutoPET dataset are utilized as unlabeled data for training models.

\subsection{Implementation Details}
All experiments are implemented in Python with PyTorch and executed on an NVIDIA A800 GPU. The student model adopts a 3D U-Net architecture as the segmentation backbone. We utilize the Stochastic Gradient Descent (SGD) optimizer with an initial learning rate of $0.01$, a weight decay of $1 \times 10^{-4}$, and a momentum of $0.9$ to update the network parameters with the maximum iteration number set to $30,000$. A polynomial learning rate schedule is adopted, where the learning rate is updated as $(1.0 - t/T)^{0.9}$, with $t$ and $T$ denote the current and maximum iteration numbers, respectively. The patch size of the student model is set to $128 \times 128 \times 128$ sub-volumes to maintain compatibility with the input dimensions of the generalist models. During the training stage, data augmentation techniques, including random cropping, flipping, and rotation, are applied to enlarge the training set and avoid over-fitting. During the inference stage, the final volumetric segmentation results are obtained using a sliding-window strategy to ensure spatial consistency.

\subsection{Experiment Results}

As shown in Fig. \ref{fig2} and Table \ref{tab:segmentation_results}, our method achieves the best performance across all semi-supervised settings. In the extreme scenario with only 5 labeled cases, our method attains a Dice score of 46.93\%, significantly outperforming the second-best method CPS and the single-modality foundation model SemiSAM+, which validates the effectiveness of dual modality mutual distillation. As the number of labeled cases increases to 10 and 20, our method maintains consistent superiority, while semi-supervised methods such as UA-MT and SASSNet exhibit limited performance gains, indicating that our framework leverages unlabeled data more effectively. Notably, our method demonstrates particularly prominent advantages in boundary accuracy metrics: the HD95 drops to 8.50 voxels in the 5 labeled setting, representing a 41\% reduction compared to SemiSAM+, benefiting from the effective fusion of anatomical priors from SAM-Med3D and SegAnyPET. Furthermore, with 10 labeled cases, our semi-supervised method surpasses the dual-modality fully-supervised baseline by approximately 9\%, demonstrating that mining unlabeled data through cross-modality consensus mechanisms can break through the performance bottleneck of limited annotations. These results consistently indicate that the mutual distillation framework successfully bridges the strengths of student models and generalist models while alleviating the annotation burden for PET/CT.

\definecolor{lightgray}{gray}{0.92}
\definecolor{highlight}{RGB}{230,240,250}
\begin{table}[t]
\centering
\caption{Ablation study on key components of our framework \underline{(5 labeled cases)}.}
\label{tab:ablation}
\resizebox{0.8\columnwidth}{!}{%
\begin{tabular}{@{}lcccc@{}}
\toprule
\rowcolor{lightgray}
\textbf{Configuration} & \textbf{Dice(\%)$\uparrow$} & \textbf{RAVD(\%)$\downarrow$} & \textbf{ASD$\downarrow$} & \textbf{HD95$\downarrow$} \\
\midrule
\multicolumn{5}{@{}l@{}}{\textit{(a) Teacher configuration}} \\
\quad SAM-Med3D only & 41.3$\pm$22.3 & 38.6$\pm$19.6 & 31.2$\pm$18.3 & 14.8$\pm$9.6 \\
\quad SegAnyPET only & 38.9$\pm$21.6 & 40.1$\pm$20.3 & 34.7$\pm$20.2 & 15.9$\pm$10.2 \\
\rowcolor{highlight}
\quad \textbf{Both} & \textbf{46.9$\pm$21.5} & \textbf{37.0$\pm$18.8} & \textbf{23.7$\pm$16.6} & \textbf{8.5$\pm$7.3} \\
\midrule
\multicolumn{5}{@{}l@{}}{\textit{(b) IoU filtering strategies}} \\
\quad w/o filtering & 44.6$\pm$21.9 & 37.9$\pm$18.9 & 26.5$\pm$16.8 & 10.6$\pm$8.3 \\
\quad Fixed threshold ($\tau$=0.5) & 45.8$\pm$22.0 & 37.2$\pm$18.6 & 25.2$\pm$16.5 & 9.3$\pm$7.9 \\
\rowcolor{highlight}
\quad \textbf{Adaptive top-50\%} & \textbf{46.9$\pm$21.5} & \textbf{37.0$\pm$18.8} & \textbf{23.7$\pm$16.6} & \textbf{8.5$\pm$7.3} \\
\midrule
\multicolumn{5}{@{}l@{}}{\textit{(c) Prompting strategies}} \\
\quad Single-point ($K$=1) & 45.2$\pm$22.0 & 37.6$\pm$18.9 & 24.9$\pm$16.2 & 9.1$\pm$7.6 \\
\rowcolor{highlight}
\quad \textbf{Multi-point ($K$=3)} & \textbf{46.9$\pm$21.5} & \textbf{37.0$\pm$18.8} & \textbf{23.7$\pm$16.6} & \textbf{8.5$\pm$7.3} \\
\bottomrule
\end{tabular}%
}
\end{table}

Table~\ref{tab:ablation} validates each component for our framework. Dual teachers outperform single-modality variants, confirming CT and PET complementarity and the efficacy of strong foundation model priors. Adaptive top 50\% IoU filtering surpasses no filtering and fixed threshold by 2.3\% and 1.1\% Dice respectively. Multi-point prompting ($K$=3) improves over single-point by 1.7\% Dice for capturing heterogeneous PET uptake.

\section{Conclusion}
This paper presents a novel mutual distillation framework leveraging dual foundation models for semi-supervised PET/CT organ segmentation. By synergistically exploiting SAM-Med3D for CT and SegAnyPET for PET, our approach achieves automatic segmentation without manual prompting. On the AutoPET multi-organ segmentation benchmark, MuDuo achieves a higher Dice score with only five labeled cases and a substantially lower HD95 than SemiSAM+, validating the efficacy of cross-modality mutual distillation between dual foundation models.

\begin{credits}
\subsubsection{\ackname} This work was supported by the ``Pioneer'' and ``Leading Goose'' R\&D Program of Zhejiang (No.~2025C02014), and the National Natural Science Foundation of China (No.~62506328). The authors also thank the High Performance Computing Center of Central South University, and the Graduate Research and Innovation Project of Central South University (No.~1053320241117).

\subsubsection{\discintname}
The authors have no competing interests to declare that are
relevant to the content of this article.
\end{credits}
\bibliographystyle{splncs04}
\bibliography{ref}
\end{document}